\documentclass{article}

\usepackage{iclr2027_conference, times}

\usepackage{amsmath,amsfonts,bm}

\def\eqref#1{equation~\ref{#1}}

\def\1{\bm{1}}

\DeclareMathAlphabet{\mathsfit}{\encodingdefault}{\sfdefault}{m}{sl}
\SetMathAlphabet{\mathsfit}{bold}{\encodingdefault}{\sfdefault}{bx}{n}

\usepackage{hyperref}
\usepackage{url}
\usepackage{svg}
\usepackage[dvipsnames,table]{xcolor}

\usepackage[T1]{fontenc}
\usepackage{amsmath,amssymb}
\usepackage{graphicx}
\usepackage{booktabs,array,tabularx}
\usepackage{multirow}

\title{
    \textsc{Möbius}: A Unified Motion-Language Model \\
    Bridging Understanding and Generation
}

\iclrfinalcopy
\author{
\begin{tabular}{@{}l@{}}
Jiazhi Shu$^{1,2,*}$,
Chengfeng Zhao$^{1,*}$,
Jinyi Wang$^{2,\dagger}$,
Yubo Zhao$^{1,2}$,
Zhiyang Dou$^{3}$ \\
Ronghui Li$^{2}$,
Yu Lu$^{2}$,
Xinwei Jiang$^{2}$,
Zijiao Zeng$^{2,\ddagger}$,
Yuan Liu$^{1,\ddagger}$
\end{tabular} \\
\\
$^{1}$The Hong Kong University of Science and Technology \\
$^{2}$Tencent Games Motus \\
$^{3}$Massachusetts Institute of Technology \\
\\
Project page: \href{https://jiazhishu.github.io/mobius/}{\textcolor{red}{\nolinkurl{https://jiazhishu.github.io/mobius/}}}
}

\begin{document}

\maketitle

\begingroup
    \renewcommand{\thefootnote}{\fnsymbol{footnote}}
    \footnotetext[1]{Equal contribution.}
    \footnotetext[2]{Project lead.}
    \footnotetext[3]{Corresponding authors.}
\endgroup

\lhead{Preprint}

\begin{figure}[h]
  \centering
  \includegraphics[width=1.03\linewidth]{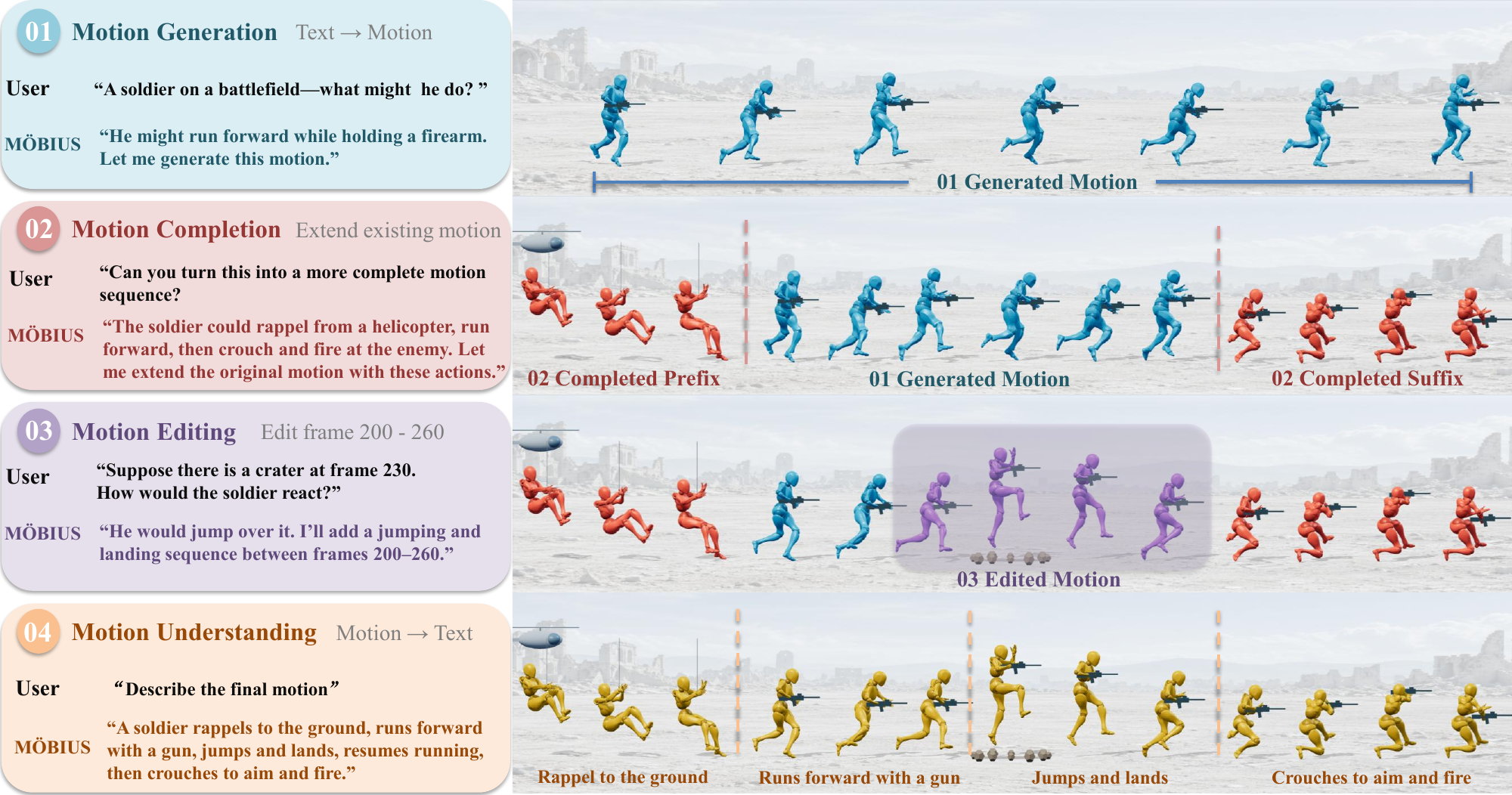}
  \vspace{-20pt}
    \caption{\textbf{Multi-turn motion creation with \textsc{Möbius}.}
    A motion sequence is progressively generated, completed, edited, and understood
    through natural language in multiple turns.}
  \label{fig:teaser}
\end{figure}

\begin{abstract}
We aim to build a truly unified motion model that can understand, generate, and edit human motion within a single framework. Existing motion UMMs remain only partially unified: motion understanding is handled by a language model, while generation is often delegated to a separate diffusion module operating over learned motion latents. We introduce \textbf{\textsc{Möbius}}, which brings the entire diffusion process into the same multimodal representation space used for motion understanding. Clean motion, noisy motion, and text jointly participate in shared attention, while role-specific parameters process understanding and generation streams. Importantly, \textsc{Möbius} operates directly on continuous frame-level raw motion, requiring neither discrete tokenization nor a pretrained motion VAE. This simple formulation naturally supports motion captioning, text-to-motion generation, temporal completion, and instruction-guided editing within a single model. On HumanML3D, \textsc{Möbius} achieves state-of-the-art performance among unified motion models. Scaling to 1.7B parameters and 1,700 hours of motion data further demonstrates that our framework is both effective and scalable.
\end{abstract}

\section{Introduction}
\label{sec:intro}
Unifying understanding and generation within a single model is a central goal of multimodal intelligence, including motion-language modeling. Existing approaches have largely followed two paradigms. Diffusion-based methods~\citep{mdm,mld,molingo,hymotion} excel at high-fidelity motion generation but typically lack native motion understanding capabilities. In contrast, language-model-based approaches~\citep{tm2t,motiongpt,motionagent} unify multiple motion–language tasks autoregressively, but struggle in generation quality. This gap motivates a unified model that excels at both understanding observed motion and generating high-fidelity motion from natural-language instructions.

\begin{figure}[t]
\centering
\includegraphics[width=1\linewidth]{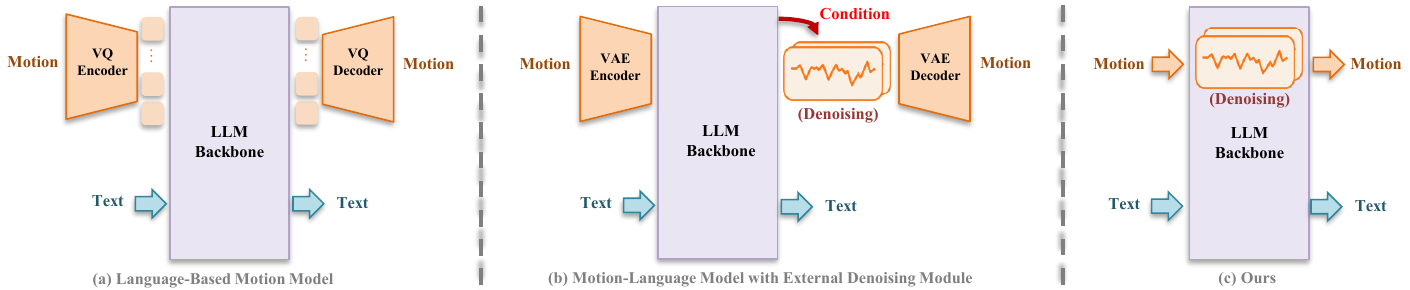}
\caption{\textbf{Architectural comparison of unified motion--language modeling paradigms.}
(a) Language-based motion models discretize motion into tokens and model motion
and text within a shared LLM.
(b) Continuous motion--language models use the LLM to produce conditioning
representations, while motion is generated by an external diffusion module.
(c) \textsc{Möbius} brings the denoising process into the unified multimodal
backbone, allowing noisy motion and language to interact throughout generation.}
\label{fig:paradigms}
\end{figure}

Recent unified motion–language models have attempted to unify motion understanding and generation~\citep{motiongpt3,llamo,unimotion_eccv} (see Fig.~\ref{fig:paradigms} for an architectural comparison). In these approaches, the language model first produces a shared multimodal representation, which is then passed as conditioning to an external diffusion model for motion generation. While this design allows understanding and generation to interact through a common feature space, the actual denoising process remains largely separated from the language model. The unified representation specifies what motion to generate, but does not directly participate in how the motion is progressively constructed. Consequently, motion understanding and motion generation are only partially unified. This motivates a more tightly integrated formulation in which not only the conditioning representation, but also the diffusion process itself operates within the unified multimodal feature space. We therefore ask: \textit{Can motion understanding and the entire motion generation process be unified within a single representation space?}

To address this question, we introduce \textsc{Möbius}, which brings the entire motion diffusion process into the same unified representation space as motion understanding. Both clean motion for understanding and noisy motion under generation are represented as motion tokens and jointly participate in the same attention with language tokens. Inspired by the Mixture-of-Transformers (MoT)~\citep{mot} design, we use dedicated parameters to process clean and noisy motion streams while allowing them to communicate through shared attention. 
As a result, motion generation is no longer performed by an external diffusion module conditioned on features produced by the language model. Instead, noisy motion is progressively denoised directly within the unified multimodal space, where it can continuously interact with clean motion and language semantics. The generated motion can in turn be fed back as clean motion tokens for subsequent understanding and reasoning, forming a fully unified loop between motion generation and understanding. This tighter integration leads to stronger motion generation, understanding, and cross-modal interaction within a unified model.

A key advantage of this formulation is that it removes the need to learn an intermediate motion latent space. Rather than relying on a pretrained motion VAE, tokenizer, or task-specific encoder~\citep{motiongpt,motiongpt3,motionstreamer,unimotion_eccv}, \textsc{Möbius} operates directly on raw motion representations. Clean motion used for understanding and noisy motion used for generation share the same representation interface: both are projected into the unified multimodal space, with their roles distinguished only by whether they represent clean observations or noisy states undergoing denoising. This makes motion understanding and generation two modes of computation over the same underlying motion representation, rather than two tasks connected through separately learned latent spaces. The resulting framework is considerably simpler, requiring no additional representation learning stage or pretrained motion encoder, while our experiments show that directly modeling raw motion is sufficient to achieve strong performance for both understanding and generation. 

To preserve the generalization ability of the pretrained LLM used in \textsc{Möbius}, we adopt a curriculum training strategy. Joint training can introduce interference between understanding and generation, as gradients from the generation objective can propagate into the understanding parameters through shared attention. We therefore first train on motion understanding tasks, allowing the LLM to acquire meaningful motion semantics, and then introduce the generation objective on top of this learned representation. This curriculum training strategy provides a stable foundation for generation while better preserving the model’s original generalization ability.

We evaluate two model variants to assess benchmark performance and the benefits of scaling model capacity and training data. Under the HumanML3D evaluation protocol~\citep{humanml3d}, \textsc{Möbius}-0.6B achieves state-of-the-art results in both text-to-motion generation and motion captioning, outperforming relevant diffusion-based, language-model-based, and motion–language baselines. \textsc{Möbius}-0.6B also demonstrates strong capabilities in motion completion, motion in-betweening, and instruction-guided editing. Controlled ablations substantiate the effectiveness of our reasoner–generator architecture, showing that coupling motion understanding and generation yields improvements over commonly used architectural alternatives. We further present \textsc{Möbius}-1.7B, a larger variant trained on 1,700 hours of high-quality motion data, which delivers additional improvements across all motion-language tasks on both controlled and uncontrolled user inputs.


\section{Related Work}
\label{sec:related}
\paragraph{Unified Multimodal Models.}
Unified multimodal models bring understanding and generation into
a shared framework, as exemplified by recent systems such as
BAGEL~\citep{bagel} and Cosmos3~\citep{cosmos3}.
Existing approaches adopt unified sequence
modeling~\citep{chameleon,emu3} or hybrid objectives combining text
autoregression with visual denoising~\citep{transfusion,showo2}.
Mixture-of-Transformers introduces modality-specific computation
with shared attention~\citep{mot}, while task-decoupled architectures
further address interference between understanding and
generation~\citep{uncovering}.
These developments motivate our investigation of specialized
computation within a unified motion--language model.

\paragraph{Human Motion Modeling.}
Human motion research spans diffusion-based generation
\citep{mdm,mld,emdm,hymotion,comovi},
discrete-token generation
\citep{t2mgpt,momask,scamo,gotozero},
retrieval and understanding
\citep{tmr,MG-MotionLLM,motionllm},
editing and completion
\citep{motionfix,simmotionedit,condmdi,pay},
and spatial control
\citep{tlcontrol,omnicontrol}.
These methods are typically developed for particular tasks, with
objectives and conditioning mechanisms tailored to their settings.
Generalist frameworks consolidate multiple motion synthesis
tasks~\citep{motionlab,genmo,umo}, motivating broader unification that
also incorporates language-based motion understanding.

\paragraph{Unified Motion Understanding and Generation.}
Motion--language models support understanding and generation through
discrete motion vocabularies~\citep{tm2t,motiongpt,motionagent,lamp,avatargpt},
with extensions to conversational and multimodal
interaction~\citep{motionchain,versatilemotion,unimumo,m3gpt}.
Continuous alternatives avoid discrete quantization:
MoTe~\citep{mote} and UniMotion~\citep{unimotion_3dv} jointly model motion and
text through latent diffusion, while MotionGPT3~\citep{motiongpt3},
LLaMo~\citep{llamo}, and UniMotion~\citep{unimotion_eccv} integrate continuous
motion representations with language-model backbones for unified understanding
and generation.
These approaches avoid discrete motion tokens but still rely on learned motion
latent spaces or encoder--decoder interfaces.
\textsc{Möbius} instead operates directly on framewise motion features and
brings denoising into the unified multimodal backbone.
Clean motion, noisy motion, and text share attention, while role-specific
projectors and Transformer parameters specialize their processing.
This allows noisy motion to continuously interact with motion and language
context throughout denoising.
\section{Method}
\label{sec:method}

\begin{figure}[t]
\vspace{-15pt}
    \centering
    \includegraphics[width=\linewidth]{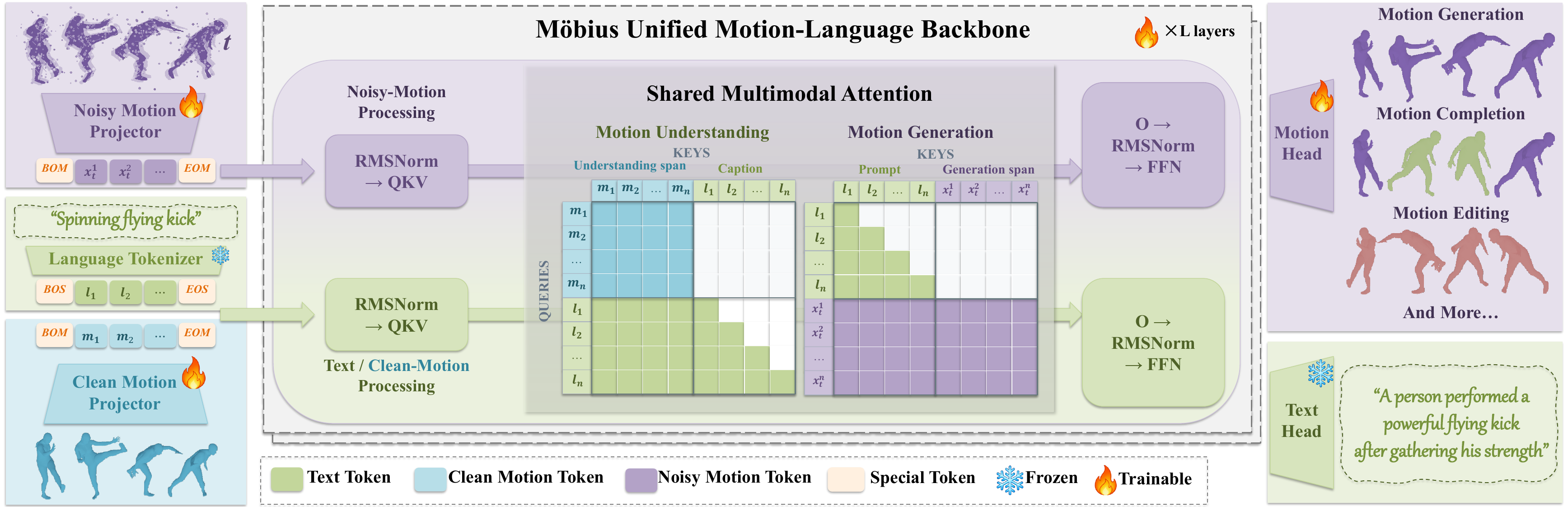}
    \vspace{-5pt}
    \caption{\textbf{Overview of \textsc{Möbius}.} Motion understanding and generation share a unified multimodal backbone. Text tokens, clean motion frames, and noisy target frames with flow time $t$ participate in shared attention at every layer, while role-specific parameters handle normalization, projections, and feed-forward transformations. The text head autoregressively predicts captions, and the motion head recovers target frames in parallel for generation, temporal completion, and editing. The central matrices illustrate attention visibility for motion understanding and generation.}
    \label{fig:architecture}
    \vspace{-10pt}
\end{figure}

\textsc{Möbius} unifies motion understanding and generation within a shared
representation space (see Figure~\ref{fig:architecture}). Starting from
frame-level raw motion (Section~\ref{sec:interface}), \textsc{Möbius} routes
clean and noisy motion tokens to role-specific parameters while allowing them
to interact through shared attention (Section~\ref{sec:branches}). It further
adopts a curriculum training strategy to establish a stable foundation for
generation while preserving the model's original generalization ability
(Section~\ref{sec:training}).

\subsection{Frame-level raw motion representation}
\label{sec:interface}

In our framework, we choose frame-level raw motion as the motion representation instead of learning any additional motion encoder.
In previous methods, discrete motion tokenizers introduce quantization error
\citep{tm2t,motiongpt,motionagent}, while VAE-based approaches
compress motion to temporally downsampled latents~\citep{motiongpt3,llamo,unimotion_eccv}, losing frame-level information. Instead, we follow HY-Motion~\citep{hymotion} to represent a motion clip as $X=[x_1,\ldots,x_T]^\top\in\mathbb{R}^{T\times D}$ with motion token dimension $D{=}201$ comprising joint positions and rotations. Special tokens $\langle\mathrm{BOM}\rangle$ and $\langle\mathrm{EOM}\rangle$ are used to delimit the motion token sequence from text. Two dedicated projectors map clean motion tokens and noisy motion tokens to the shared representation space for understanding and generation, and corresponding motion heads decode it back to the raw motion space. In comparison with existing methods~\citep{motiongpt,llamo,motiongpt3,unimotion_eccv} that learn different encoders for understanding and generation, we show that the raw motion representation can achieve strong performance in a unified representation space.

\subsection{Unified feature space attention with MoT}
\label{sec:branches}

To process all the clean or noisy raw motion and text prompts, we adopt a Mixture-of-Transformers~\citep{mot} structure. In MoT, tokens are routed to different parameters for normalization and linear projection but are then processed by a unified attention operation. We introduce how we route different tokens, and then we discuss how to conduct generative denoising.

\paragraph{Routing scheme.} We route tokens based on their roles. Text tokens, boundary tokens, and clean motions are intended for understanding and are processed by the understanding parameters. Noisy target motions are used for generation and thus processed by the generation parameters
\begin{equation}
r_i =
\begin{cases}
\mathrm{gen}, & i \text{-th token is a noisy motion token},\\
\mathrm{und}, & \text{otherwise},
\end{cases}
\label{eq:routing}
\end{equation}
where $r_i$ determines the set of parameters used in normalization and linear projection (``to QKV" in standard attention layers). After that, all projected tokens are processed by a shared attention operation to exchange information in the same feature space.

\paragraph{Bidirectional attention for motion.}
In our design, motion token attention follows a different bidirectional visibility pattern from the causal attention of autoregressive text tokens. Specifically, we allow all motion tokens within the same motion sequence to attend bidirectionally to one another, regardless of whether they represent clean motion for understanding or noisy motion for generation. For understanding, the entire observed motion is already available as input, so bidirectional attention enables each frame to capture context from the complete motion sequence. For generation, all noisy target frames are instantiated jointly at each denoising step, and bidirectional attention allows each frame to exploit both past and future temporal context during refinement. Text tokens, in contrast, retain the causal attention pattern of the pretrained language model. In implementation, the target generated motion sequence is always put at the end of the whole sequence with its generation sequence length specified before denoising, allowing the full noisy sequence to be progressively refined as a whole.

\subsection{Curriculum training strategy}
\label{sec:training}
We train \textsc{Möbius} in two stages. Directly optimizing motion understanding and generation together is unstable, as the randomly initialized generation pathway can disrupt the pretrained language representations and degrade the model's understanding ability. We therefore adopt an understanding-first training strategy: the model first learns to interpret motion in the language space, and generation is introduced only after this representation has been established.

\paragraph{Phase I: motion understanding.}

We initialize our model from Qwen3~\citep{qwen3} and train it together with the motion projector on motion captioning. Given a clean motion sequence $X$, an instruction $C$, and a target response $Y=(y_1,\ldots,y_L)$, we use the standard autoregressive language modeling objective:
\begin{equation}
\mathcal{L}_{\mathrm{und}}
=
-\mathbb{E}_{(X,C,Y)}
\left[
\frac{1}{L}
\sum_{j=1}^{L}
\log p_{\theta_{\mathrm{und}}}
(y_j\mid y_{<j},C,X)
\right].
\label{eq:m2t_loss}
\end{equation}
This stage allows the pretrained LLM to acquire motion semantics while preserving its original language reasoning capability.

\paragraph{Phase II: motion generation.}
We then initialize the generation parameters from the trained understanding parameters and freeze the understanding parameters. For a target motion $X_{\mathrm{gt}}$, we sample $\epsilon\sim\mathcal{N}(0,I)$ and $t\sim\mathcal{U}[0,1]$, and construct a noisy motion
\begin{equation}
X_t=(1-t)\epsilon+tX_{\mathrm{gt}}.
\label{eq:flow}
\end{equation}
The model is trained to iteratively denoise $X_t$ to recover the clean motion while attending to the frozen text and motion representations produced by the understanding part:
\begin{equation}
\mathcal{L}_{\mathrm{gen}}
=
\mathbb{E}
\left[
\left\|
f_{\theta_{\mathrm{gen}}}
(X_t,t;H_{\theta_{\mathrm{und}}}(C))
-
X_{\mathrm{gt}}
\right\|_F^2
\right].
\label{eq:gen_loss}
\end{equation}
This staged training lets generation build on an already learned motion--language representation, avoiding interference with the pretrained understanding capability.

\paragraph{Unified motion editing and completion.} The same framework naturally extends to a broad range of motion editing and completion tasks by treating the source motion as part of the conditioning context and generating the desired target motion. We also train our model on these tasks. Depending on which parts of the source are observed, the model supports temporal completion, including prediction, backcasting, in-betweening and keyframe infilling. It also supports semantic editing, where a source motion is modified according to a natural-language instruction. In all cases, the observed source motion is encoded as clean context, while the generation process predicts the missing or edited target motion. We construct diverse training pairs covering these settings and jointly train a unified editing model without introducing task-specific architectures.
\section{Experiments}
\label{sec:experiments}

\subsection{Experimental Setup}
\label{sec:exp-setup}

We evaluate \textsc{Möbius} on text-to-motion generation, motion captioning,
instruction-guided editing, and temporal completion tasks, and conduct controlled
ablations to examine its key design choices. \textbf{Bold} and \underline{underlined} metric numbers indicate the best and second-best performance respectively.

\paragraph{Text-to-motion generation and motion captioning.} We consider two model scales: \textsc{Möbius}-0.6B and \textsc{Möbius}-1.7B. The standard 0.6B variant is trained exclusively on HumanML3D~\citep{humanml3d}, while the 1.7B variant additionally uses an
internal training corpus containing 1,700 hours of annotated motion. For the generation task, we use the MS-272 evaluator introduced by MotionStreamer~\citep{motionstreamer} to compare against representative baselines from diffusion-based models~\citep{mdm,hymotion,motionstreamer}, autoregressive models~\citep{t2mgpt,gotozero,momask}, and unified motion-language models~\citep{motiongpt,motiongpt3}. For motion captioning, we follow the TM2T protocol~\citep{tm2t} to compare with those methods that natively support understanding~\citep{tm2t,motiongpt,motiongpt3,unimumo}.

\paragraph{Motion editing and completion.} For editing and completion tasks, we report the performance on MotionLab~\citep{motionlab} benchmark of both \textsc{Möbius}-0.6B trained on HumanML3D~\citep{humanml3d} and an additional variant that incorporates MotionFix~\citep{motionfix} data in training. The compared baselines include the task-specific baseline MotionFix~\citep{motionfix}, the generalist model
MotionLab~\citep{motionlab}, and also the unedited source motion.

\begin{figure}[t]
    \vspace{-15pt}
  \centering
  \includegraphics[width=\textwidth]{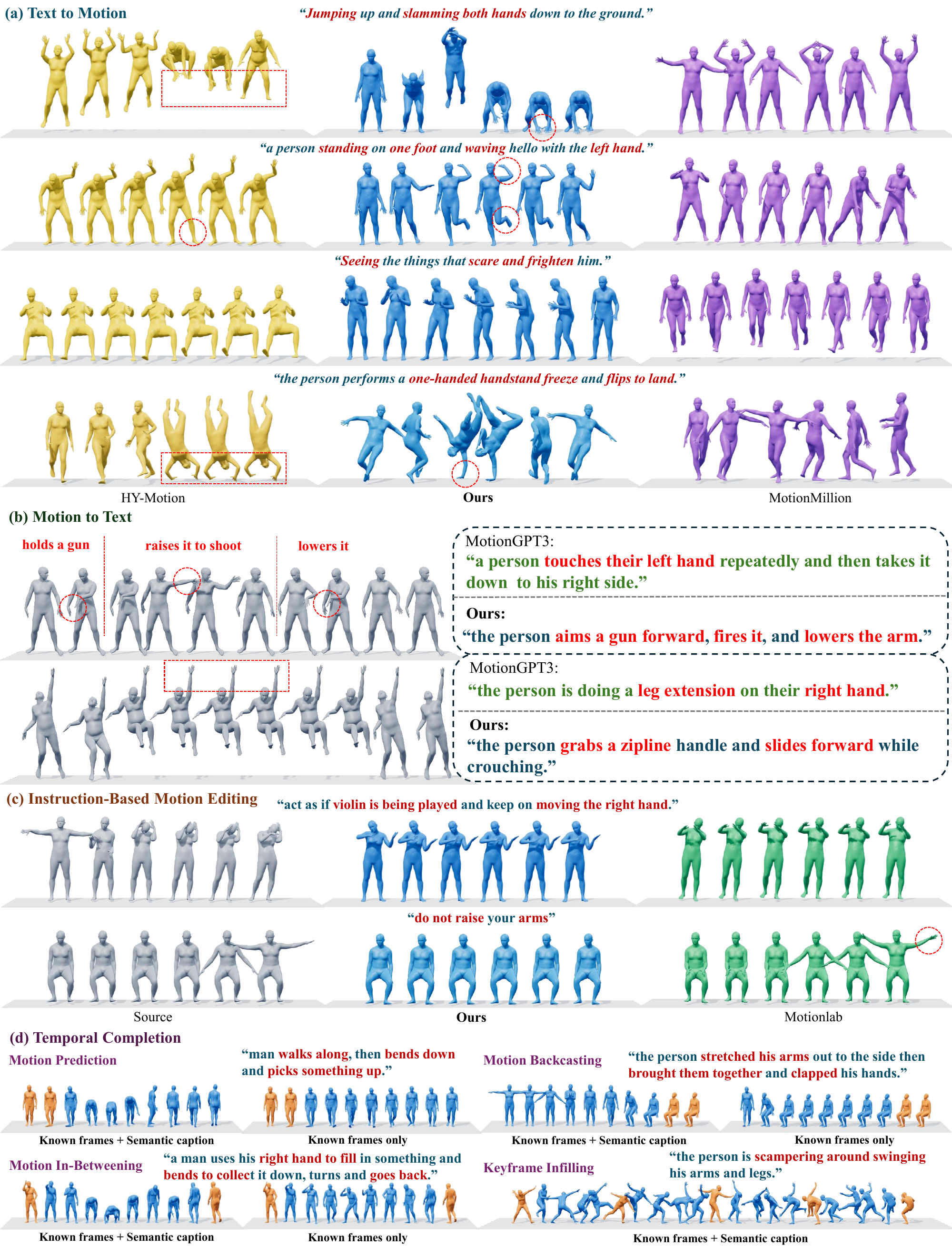}
  \caption{ 
  \textbf{Qualitative results on motion understanding and generation.} (a) Text-to-motion generation results of HY-Motion~\citep{hymotion} (yellow), \textsc{Möbius} (blue), and MotionMillion~\citep{gotozero} (purple). (b) Motion-to-text understanding compared with MotionGPT3~\citep{motiongpt3}. (c) Instruction-based editing: source (gray), \textsc{Möbius} (blue), and MotionLab~\citep{motionlab} (green). (d) Temporal completion: prediction, backcasting, in-betweening, and keyframe infilling; known and generated frames are orange and blue, respectively.
  }
  \label{fig:qualitative}
    \vspace{-10pt}
\end{figure}

\subsection{Quantitative Evaluation}
\label{sec:quantitative}

\begin{table}[t]
\vspace{-15pt}
\caption{\textbf{Comparison of text-to-motion generation on HumanML3D~\citep{humanml3d}.}
``Task" indicates whether a model is generation-only (Gen.) or unified for
motion understanding and generation (Uni.). We use the
evaluator from MotionStreamer~\citep{motionstreamer}.}
\label{tab:t2m}
\centering\scriptsize
\setlength{\tabcolsep}{2.5pt}
\begin{tabularx}{\linewidth}{
@{}l
c
*{5}{>{\hsize=.94\hsize\centering\arraybackslash}X}
>{\hsize=1.18\hsize\centering\arraybackslash}X@{}}
\toprule
Method & Task
& R@1$\uparrow$
& R@2$\uparrow$
& R@3$\uparrow$
& FID$\downarrow$
& MMDist$\downarrow$
& Diversity$\rightarrow$\\
\midrule

GT & -- & 0.7703 & 0.9030 & 0.9442 & 0.0000 & 14.8785 & 27.7705\\

\midrule
\multicolumn{8}{l}{\cellcolor{lightgray!30}\textit{Trained on HumanML3D}}\\

T2M-GPT~\citep{t2mgpt}
& Gen.
& 0.5516 & 0.7056 & 0.7788
& 25.4913 & 19.0912 & 25.5949\\

MoMask~\citep{momask}
& Gen.
& 0.6404 & 0.7974 & 0.8609
& 21.0729 & 18.1216 & 25.9789\\

MDM~\citep{mdm}
& Gen.
& 0.5208 & 0.6937 & 0.7701
& 35.5169 & 19.4246 & 25.3383\\

MotionStreamer~\citep{motionstreamer}
& Gen.
& 0.6303 & 0.7865 & 0.8498
& \underline{12.2110} & \underline{16.5810} & \underline{27.4637}\\

MotionGPT~\citep{motiongpt}
& Uni.
& 0.4940 & 0.6352 & 0.6944
& 23.6811 & 19.6781 & 25.5410\\

MotionGPT3~\citep{motiongpt3}
& Uni.
& \underline{0.6709} & \underline{0.8242} & \underline{0.8817}
& 20.9913 & 17.5664 & 25.6889\\

Ours (0.6B)
& Uni.
& \textbf{0.8209} & \textbf{0.9397} & \textbf{0.9670}
& \textbf{6.1291} & \textbf{13.9973} & \textbf{27.6107}\\

\midrule
\multicolumn{8}{l}{\cellcolor{lightgray!30}\textit{Trained on large-scale dataset}}\\

Go to Zero (3B)~\citep{gotozero}
& Gen.
& 0.7401 & 0.8772 & 0.9229
& 23.7550 & 15.3806 & 26.6340\\

HY-Motion (1B)~\citep{hymotion}
& Gen.
& \underline{0.7899} & \underline{0.9177} & \underline{0.9516}
& \underline{13.5194} & \underline{14.7748} & \underline{27.4183}\\

Ours (1.7B)
& Uni.
& \textbf{0.7968} & \textbf{0.9226} & \textbf{0.9524}
& \textbf{10.6642} & \textbf{14.7441} & \textbf{27.4649}\\

\bottomrule
\end{tabularx}
\vspace{-10pt}
\end{table}
\begin{table}[t]
\caption{\textbf{Comparison of motion captioning on HumanML3D~\citep{humanml3d}.}
We follow the TM2T evaluation protocol~\citep{tm2t} using raw captions.}
\label{tab:m2t}
\centering\scriptsize\setlength{\tabcolsep}{2pt}
\begin{tabularx}{\linewidth}{@{}l*{2}{>{\hsize=.95\hsize\centering\arraybackslash}X}>{\hsize=1.4\hsize\centering\arraybackslash}X*{6}{>{\hsize=.95\hsize\centering\arraybackslash}X}@{}}
\toprule
Method & BLEU-1$\uparrow$ & BLEU-4$\uparrow$ & ROUGE-L$\uparrow$ & CIDEr$\uparrow$ & BERT$\uparrow$ & R@1$\uparrow$ & R@2$\uparrow$ & R@3$\uparrow$ & MMDist$\downarrow$\\
\midrule
GT & 1.0000 & 1.0000 & 1.0000 & 4.5165 & 1.0000 & 0.5068 & 0.7064 & 0.8041 & 2.9784\\
\midrule
TM2T~\citep{tm2t} & 0.4912& 0.0760 & 0.3808 & 0.1719 & 0.3215 & 0.5180& 0.7178& 0.8079& 2.9584\\
MotionGPT~\citep{motiongpt} & 0.4605 & 0.1210 & 0.3849& 0.3409& 0.3388 & 0.5087 & 0.6880 & 0.7776 & 3.1121\\
MotionGPT3~\citep{motiongpt3} & 0.5383& 0.1605& 0.4165& 0.4077& 0.3425& 0.5333& 0.7178& 0.8043& 2.8533\\
UniMuMo~\citep{unimumo} & 0.4488 & 0.1271& 0.3560 & 0.3114 & 0.4517& 0.5162 & 0.7032& 0.7984 & 2.9658\\
\midrule
Ours 0.6B (ML3D only) & \underline{0.5664} & \underline{0.2032} & \underline{0.4636} & \underline{0.5754} & \underline{0.4541} & \underline{0.6056} & \underline{0.7936} & \underline{0.8761} & \underline{2.4693}\\
Ours 1.7B (large-scale) & \textbf{0.5874} & \textbf{0.2105} & \textbf{0.4725} & \textbf{0.5836} & \textbf{0.4657} & \textbf{0.6152} & \textbf{0.8086} & \textbf{0.8862} & \textbf{2.4143}\\
\bottomrule\end{tabularx}
\end{table}

\begin{table}[t]
\vspace{-10pt}
\caption{\textbf{Instruction-based motion editing results on MotionFix~\citep{motionfix}.}
$^\dagger$ denotes our editing variant trained with additional MotionFix pairs
in Phase II. The no-edit source and GT are shown as references and excluded
from ranking.}
\label{tab:motion-edit}
\centering
\scriptsize
\setlength{\tabcolsep}{3.2pt}
\begin{tabularx}{\linewidth}{@{}l*{7}{>{\centering\arraybackslash}X}@{}}
\toprule
Method
& FID$\downarrow$
& R@1$\uparrow$
& R@2$\uparrow$
& R@3$\uparrow$
& AvgR$\downarrow$
& MMDist$\downarrow$
& Diversity$\rightarrow$ \\
\midrule
GT
& 0.0000 & 1.0000 & 1.0000 & 1.0000 & 1.0000 & 0.0000 & 8.3500 \\
\midrule
Source motion (no edit)
& 0.1066 & 0.4640 & 0.6525 & 0.7502 & 3.5100 & 2.2817 & 8.2000 \\
MotionFix (TMED)~\citep{motionfix}
& 0.5282 & 0.3248 & 0.4877 & 0.5982 & 5.3600 & 3.1974 & 7.8900 \\
MotionLab~\citep{motionlab}
& \underline{0.1339} & \underline{0.4610} & \underline{0.6308}
& \underline{0.7315} & \underline{3.7500} & \underline{2.4112}
& \underline{8.1800} \\
\midrule
Ours (0.6B)$^\dagger$
& \textbf{0.0955} & \textbf{0.4936} & \textbf{0.6821} & \textbf{0.7828}
& \textbf{2.8026} & \textbf{1.8749} & \textbf{8.4468} \\
\bottomrule
\end{tabularx}
\vspace{-10pt}
\end{table}

\paragraph{Results.}
As shown in Table~\ref{tab:t2m}, \textsc{Möbius}-0.6B outperforms both generation-specialized and unified baselines trained on HumanML3D. The scaled \textsc{Möbius}-1.7B also achieves strong performance against large-scale motion models, attaining the best R-Precision, FID, and MMDist among the compared methods. For motion understanding, \textsc{Möbius} consistently outperforms existing unified motion-language models across all reported metrics (Table~\ref{tab:m2t}). Furthermore, \textsc{Möbius} surpasses editing-specialized baselines and even the unedited source motion (Table~\ref{tab:motion-edit}), demonstrating strong instruction-following capabilities.

\paragraph{Analysis.} 
We attribute our advantages in motion generation to the unified architecture. The routing scheme brings the entire motion diffusion process into the shared representation space, allowing the semantics of clean motion and language to continuously guide generation. Meanwhile, the curriculum training strategy helps preserve the reasoning ability inherited from the pretrained LLM, resulting in strong motion-understanding performance. Together with the frame-level raw motion representation, these designs also enable framewise control and editing that follow language instructions well, improving performance across motion–language tasks without sacrificing one capability for another.

\subsection{Qualitative Evaluation}
\label{sec:qualitative}
Figure~\ref{fig:qualitative} illustrates the versatility of \textsc{Möbius}
across both motion understanding and generation tasks. For text-to-motion
generation, \textsc{Möbius} follows challenging multi-part prompts while
preserving their temporal ordering. For motion-to-text captioning, it captures
the key actions and their temporal progression more accurately. For
instruction-guided editing, it modifies the target motion while preserving the
surrounding source content. Temporal completion requires no task-specific
architecture: changing which frames are treated as observed is sufficient to
support motion prediction, backcasting, in-betweening, and keyframe infilling.
Across these settings, the same frame-level representation supports both motion
understanding and flexible conditional generation.

\begin{table}[t]
\vspace{-10pt}
\caption{\textbf{Ablation study.}
We study motion encoding, routing, and understanding--generation training.
$^\dagger$ allows denoising to access only the final-layer context representations;
$^\ddagger$ randomly initializes the denoising parameters instead of transferring
those learned during understanding.
Their M2T scores equal those of the full model and are not ranked.
Bold and underline denote the best and second-best variants, excluding GT.}
\label{tab:ablation}
\vspace{5pt}
\centering
\scriptsize
\setlength{\tabcolsep}{2.5pt}
\begin{tabular*}{\linewidth}{@{\extracolsep{\fill}}clll*{8}{c}@{}}
\toprule
& & & &
\multicolumn{4}{c}{\textbf{Understanding (M2T)}}
& \multicolumn{4}{c}{\textbf{Generation (T2M)}} \\
\cmidrule(lr){5-8}
\cmidrule(lr){9-12}
\textbf{ID} & \textbf{Routing} & \textbf{Encoding} & \textbf{Training}
& R@1$\uparrow$
& BLEU-1$\uparrow$
& CIDEr$\uparrow$
& BERT$\uparrow$
& R@1$\uparrow$
& R@3$\uparrow$
& FID$\downarrow$
& MMDist$\downarrow$ \\
\midrule
\multicolumn{4}{@{}l}{GT}
& 0.5068 & 1.0000 & 4.5165 & 1.0000
& 0.7703 & 0.9442 & 0.0000 & 14.8785 \\
\midrule
1 & Dense & VQ & Und.$\to$Gen.
& 0.5695 & 0.5593 & 0.4568 & 0.4296
& 0.5627 & 0.7741 & 6.0912 & 16.4993 \\
2 & Role & VAE & Und.$\to$Gen.
& \underline{0.5846} & \underline{0.5619}
& \underline{0.5522} & \underline{0.4404}
& \textbf{0.8251} & \textbf{0.9655}
& 5.9970 & \textbf{13.8139} \\
\midrule
3 & Modality & Frame & Und.\ only
& 0.5675 & 0.5222 & 0.4525 & 0.4134
& -- & -- & -- & -- \\
4 & Modality & Frame & Und.$\to$Gen.
& 0.0328 & 0.0776 & 0.0040 & -0.3768
& 0.7512 & 0.9250 & 18.0517 & 15.3010 \\
5 & Modality & Frame & Gen.$\to$Align$\to$Joint
& 0.5390 & 0.5090 & 0.4100 & 0.3951
& 0.7075 & 0.8948 & 26.2865 & 16.1071 \\
\midrule
6 & Role & Frame & Gen.\ only
& -- & -- & -- & --
& 0.7877 & 0.9486 & 6.0073 & 14.2763 \\
7 & Role & Frame & Und.$+$Gen.
& 0.5291 & 0.4941 & 0.3929 & 0.3801
& 0.8125 & 0.9506 & 8.0030 & 14.1169 \\
8 & Role & Frame & Und.$\to$Gen.$^\dagger$
& 0.6001 & 0.5670 & 0.5629 & 0.4507
& 0.7621 & 0.9387 & \underline{5.7644} & 14.5591 \\
9 & Role & Frame & Und.$\to$Gen.$^\ddagger$
& 0.6001 & 0.5670 & 0.5629 & 0.4507
& 0.7721 & 0.9422 & 5.8772 & 14.5665 \\
\midrule
10 & Role & Frame & Und.$\to$Gen.
& \textbf{0.6001} & \textbf{0.5670}
& \textbf{0.5629} & \textbf{0.4507}
& \underline{0.8207} & \underline{0.9571}
& \textbf{4.8772} & \underline{14.0269} \\
\bottomrule
\end{tabular*}
\vspace{-5pt}
\end{table}
\subsection{Ablation Studies}
\label{sec:ablation}

We ablate the motion encoding, routing strategy, and understanding--generation
training in Table~\ref{tab:ablation}. All variants use Qwen3-0.6B, are trained
on HumanML3D using only the T2M and M2T objectives, and use MS-272 from
MotionStreamer~\citep{motionstreamer} as the underlying motion feature format.
For motion encoding, VQ discretizes motion using a tokenizer trained following
MotionGPT~\citep{motiongpt}, VAE maps motion into the pretrained latent space of
MotionStreamer, and Frame directly models framewise MS-272 features without an
intermediate motion codec. For routing, Dense uses a single parameter set for
all tokens, Modality separates text and motion computation but uses the same
motion parameters for both clean observations and noisy denoising states, while
Role (ours) assigns role-specific parameters to text/clean motion and noisy
motion while retaining shared attention. For training, Gen.\ only trains
generation without understanding pretraining, Und.$\to$Gen. first establishes
motion--language understanding and then transfers it to generation,
Gen.$\to$Align$\to$Joint follows the MotionGPT3~\citep{motiongpt3} schedule,
and Und.$+$Gen. jointly optimizes both objectives.

\textbf{Which motion encoding best supports unified modeling?}
We compare three representative motion interfaces in IDs 1, 2, and 10.
Since discrete VQ tokens do not naturally represent intermediate continuous
denoising states, the VQ variant uses dense computation, whereas the VAE and
frame-level variants support role-specific processing of clean and noisy motion.
VQ substantially weakens generation, while VAE remains competitive but yields
weaker motion understanding than direct frame-level encoding. Overall,
frame-level modeling provides the strongest balance between the two
capabilities. Moreover, its one-to-one correspondence between motion frames and
model positions naturally supports temporal completion and frame-conditioned
manipulation.

\textbf{Should computation be separated by modality or by role?}
With modality routing, the understanding-only model (ID 3) performs normally,
but using the same motion parameters for both clean observations and noisy
denoising states (ID 4) causes understanding to collapse. The
generate--align--joint schedule (ID 5) largely recovers understanding, but at a
substantial cost to generation quality. In contrast, separating clean and noisy
motion by computational role (ID 10) maintains strong performance in both
directions, indicating that the two roles benefit from specialized parameters
while still interacting through shared attention.

\textbf{How should understanding be transferred to generation?}
IDs 6--10 further isolate how learned motion--language understanding benefits
generation. Compared with generation-only training (ID 6), the full model
(ID 10) shows that establishing motion--language understanding first provides a
stronger foundation for synthesis. Joint optimization (ID 7) notably degrades
understanding, revealing interference between the two objectives when optimized
together. Restricting denoising to only the final-layer context representations
(ID 8) also weakens generation, demonstrating the benefit of layer-wise
interaction throughout the denoising process. Randomly initializing the
denoising parameters (ID 9) further degrades performance relative to
initialization from the parameters learned during understanding. Together,
these results support our understanding-first strategy: first establish
motion--language alignment, then transfer it to generation through parameter
initialization and layer-wise interaction while keeping the learned
understanding parameters fixed.
\section{Discussion and Conclusion}
\label{sec:conclusion}

We presented \textsc{Möbius}, a unified motion--language model that organizes computation by role rather than modality. Its role-routed architecture and continuous frame-level motion interface support motion understanding, generation, editing, and temporal completion within a single unified framework. The results also highlight the benefit of transferring learned motion understanding to generation. Experiments on HumanML3D show strong T2M and M2T performance, while scaling to 1.7B parameters and 1,700 hours of motion data further demonstrates the effectiveness of the design at larger scale.

More broadly, \textsc{Möbius} demonstrates that understanding and continuous motion generation need not be connected through a separate latent generator, but can instead operate together within a shared multimodal representation space. Our current experiments focus on controlled single-person motion settings. Future work will extend \textsc{Möbius} toward multi-person interaction, more detailed hand and finger motion, retargeting across heterogeneous body structures, and deployment in embodied systems such as real-world robots.

\subsection*{AI use statement}
Generative AI tools were used only to improve the readability and language of the manuscript, including grammar, phrasing, and stylistic editing. They were not used to develop the research methodology, design experiments, implement methods, generate or modify data, formulate mathematical claims, or interpret experimental results. The authors reviewed all AI-assisted edits and take responsibility for the final content of the paper.

\label{maintext:end}

\clearpage

\bibliography{main}
\bibliographystyle{iclr2027_conference}

\clearpage

\appendix
\providecommand{\supptodo}[1]{\textbf{[TBD: #1]}}
\providecommand{\suppfig}[2]{%
  \begin{center}\fbox{\parbox[c][#1][c]{0.94\linewidth}{\centering\small #2}}\end{center}}
\setcounter{figure}{0}
\setcounter{table}{0}
\renewcommand{\thefigure}{S\arabic{figure}}
\renewcommand{\thetable}{S\arabic{table}}

\section{Experimental Details and Inference}
\label{app:training}\label{sec:optimization}\label{app:data_analysis}
\label{app:implementation}
\label{app:architecture}
This supplement provides experimental and inference details, evaluation
protocols, and additional results.

\subsection{Datasets}
\textsc{Möbius}-0.6B uses HumanML3D motion--text data, while the 1.7B model
additionally uses an internal corpus of approximately 1,700 hours. The editing
variant adds MotionFix pairs during Phase II. Pure-text examples are
interleaved during Phase I; ``HumanML3D-only'' refers to the source of
motion--text supervision, not the absence of language pretraining or replay.


\subsection{Continuous Motion Interface}
We use 201-dimensional HY-Motion features at 30 FPS: pelvis translation (3),
root orientation in a continuous 6D representation (6), local body rotations
(126), and root-relative joint positions (66). Each frame occupies one
Transformer position. Let $\mu,\sigma\in\mathbb{R}^{201}$ denote the
training-set feature statistics, and define the elementwise scale
$\tilde\sigma=\max(\sigma,\varepsilon)$ for $\varepsilon>0$.
Normalization and its inverse are
\begin{equation}
 \bar x_i=(x_i-\mu)\oslash\tilde\sigma,
 \qquad x_i=\bar x_i\odot\tilde\sigma+\mu,
 \label{eq:supp_normalization}
\end{equation}
where operations are elementwise. All flow equations in the paper operate in this
normalized space. 

For observed motion, $P_{\mathrm{und}}$ denotes the framewise mapping
before output normalization and learned scaling:
\begin{equation}
 z_i=P_{\mathrm{und}}(\bar x_i),\qquad
 h_i^{\mathrm{obs}}=m_i\,s\frac{z_i}{\max(\|z_i\|_2,\varepsilon)}
 +(1-m_i)e_{[\mathrm{M}]},\qquad
 s=\min(\exp(b),s_{\max}).
 \label{eq:supp_observed}
\end{equation}
Here $m_i\in\{0,1\}$ indicates whether a source frame is visible, $b$ is learned, and
$e_{[\mathrm{M}]}$ represents an unrevealed frame. Noisy targets use
$h_i^{\mathrm{tgt}}=P_{\mathrm{gen}}(X_{t,i})+e(t)$ and are decoded by
$\widehat X_i=P_{\mathrm{out}}(g_i)$, where $g_i$ is the final-layer hidden state at target position $i$,
computed using the generation parameters.
The projector architectures, scale cap,
and numerical constants are recorded in Table~\ref{tab:supp_arch}.

\begin{table}[t]
\centering\small
\caption{\textbf{Architecture specification of \textsc{Möbius}}.}
\label{tab:supp_arch}
\begin{tabular}{lcc}
\toprule
Component & \textsc{Möbius}-0.6B & \textsc{Möbius}-1.7B\\
\midrule
Backbone & Qwen3-0.6B & Qwen3-1.7B\\
Vocabulary size & 151,673 & 151,673\\
Total parameters & 1.047B & 3.148B\\
Motion representation & HY201, 30 FPS & HY201, 30 FPS\\
Transformer layers & 28 & 28\\
Hidden dimension & 1,024 & 2,048\\
Query heads & 16 & 16\\
KV heads & 8 & 8\\
Head dimension & 128 & 128\\
FFN intermediate dimension & 3,072 & 6,144\\
Activation in projectors & SiLU & SiLU\\
\bottomrule
\end{tabular}
\end{table}

\begin{table*}[t]
\centering\small
\caption{\textbf{Training and inference hyper-parameters of \textsc{Möbius}-1.7B.}
We first train motion-to-text (M2T) for 20,000 steps, followed by
200,000 steps of generation-only fine-tuning.
Both stages use 2,000 warmup steps.
Training batches are dynamically packed by token count.
Inference uses EMA weights.}
\label{tab:supp_hyperparams}
\begin{tabularx}{\textwidth}{
@{}l>{\raggedright\arraybackslash}X
@{\hspace{2em}}l>{\raggedright\arraybackslash}X@{}}
\toprule
Hyper-parameter & Value & Hyper-parameter & Value\\
\midrule
\multicolumn{2}{@{}l}{\textit{Training schedule}}
& \multicolumn{2}{l@{}}{\textit{Generation objective}}\\
M2T training steps & 20,000
& Prediction target & $x_0$\\
Generation training steps & 200,000
& Timestep sampling & Uniform\\
Warmup steps (each stage) & 2,000
& Motion dimension & 201\\
Token budget / GPU & 65,536
& Text condition dropout & 0.1\\
Gradient accumulation & 1
& Geometric loss & Enabled\\
\midrule
\multicolumn{2}{@{}l}{\textit{Generation optimization}}
& \multicolumn{2}{l@{}}{\textit{Inference}}\\
Optimizer & AdamW
& Sampler & Euler (rectified flow)\\
Learning rate & $10^{-4}$
& Sampling steps & 50\\
Weight decay & 0
& CFG scale & 3.0\\
AdamW $(\beta_1,\beta_2)$ & $(0.9,\,0.95)$
& Timestep shift & 1.0\\
AdamW $\epsilon$ & $10^{-8}$
& Model weights & EMA\\
LR scheduler & Constant with warmup
& \multicolumn{2}{l@{}}{\textit{Runtime}}\\
Max gradient norm & 1.0
& Training strategy & FSDP\\
EMA decay & 0.999
& Gradient checkpointing & Enabled\\
Trainable modules & Generation experts and interfaces
& CPU offload & Disabled\\
Precision & bf16
& Final GPU count & 16\\
\bottomrule
\end{tabularx}
\end{table*}

\subsection{Task Construction}
\label{app:task_layouts}
Let $C$ be the text instruction, $B_S$ an observed-motion span, $B_t$ the
noisy target span, and $Y$ a text response. The mask $m$ controls which source
frames are provided as clean context to the understanding parameters. Task-specific context is
summarized in Table~\ref{tab:supp_layouts}.

\begin{table}[t]
\centering\small
\caption{\textbf{Unified task construction.} $\Omega$ denotes observed temporal indices.
For temporal tasks, observations use the target timeline.}
\label{tab:supp_layouts}
\begin{tabularx}{\linewidth}{@{}llX@{}}\toprule
Task & Context / output & Observed motion\\\midrule
T2M & $C\rightarrow B_t$ & None\\
M2T & $[C,B_S]\rightarrow Y$ & Full source\\
Editing & $[C,B_S]\rightarrow B_t$ & Full source\\
Prediction & $[C,B_S]\rightarrow B_t$ & Prefix $\Omega$\\
Backcasting & $[C,B_S]\rightarrow B_t$ & Suffix $\Omega$\\
In-betweening & $[C,B_S]\rightarrow B_t$ & Prefix and suffix\\
Keyframe infilling & $[C,B_S]\rightarrow B_t$ & Sparse frames\\
\bottomrule\end{tabularx}
\end{table}

For editing, the source is supplied only as understanding context; the target
starts from Gaussian noise, as in T2M. Temporal tasks also provide observed
frames through the understanding context, without pinning generated frames
after sampling updates. Their observation masks and initialization settings
are specified with the evaluation protocol below.

\subsection{Inference}
\label{app:sampling}
The model predicts a clean endpoint using the generation parameters. Text-only classifier-free
guidance retains the same motion context in both passes. Here
$f_\theta(X_t,t;C,S,m)$ abbreviates the endpoint predictor with context
encoded using the understanding parameters, and
$\theta=(\theta_{\mathrm{und}},\theta_{\mathrm{gen}})$:
\begin{equation}
 \widehat X_g(X_t,t)=f_\theta(X_t,t;\varnothing,S,m)
 +w\bigl[f_\theta(X_t,t;C,S,m)-f_\theta(X_t,t;\varnothing,S,m)\bigr].
 \label{eq:supp_cfg}
\end{equation}
The source is absent for T2M. For a grid $0=t_0<\cdots<t_N=1$, Euler sampling
converts the predicted endpoint to a velocity:
\begin{equation}
 X_{t_{n+1}}=X_{t_n}+(t_{n+1}-t_n)
 \frac{\widehat X_g(X_{t_n},t_n)-X_{t_n}}{\max(1-t_n,\varepsilon)}.
 \label{eq:supp_euler}
\end{equation}
The network is evaluated at $t_n<1$, and the final motion is denormalized.
Because context cannot attend to noisy targets, its layer-wise keys and
values can be reused across sampling steps; conditional and null-text
passes use separate caches.

For M2T, the model uses the understanding parameters to generate
$p(Y\mid C,S)=\prod_j p(y_j\mid y_{<j},C,S)$ autoregressively.
Pure-text inference uses the same parameters without a motion span.
Sampler and text-decoding settings are listed in Table~\ref{tab:supp_inference}.

\begin{table}[t]
\centering\small
\caption{\textbf{T2M and M2T inference-pipeline settings.} The response-token limit
is the M2T script default. Editing and temporal-inpainting conditions are
specified separately in their task protocols.}
\label{tab:supp_inference}
\begin{tabularx}{\linewidth}{@{}lXX@{}}
\toprule
Setting & Text-to-motion & Motion-to-text\\
\midrule
Sampler / decoder
& Euler with clean-endpoint-to-velocity conversion
& Autoregressive greedy decoding\\
Steps / time grid
& 50; uniform on $[0,1]$; timestep shift $1.0$
& Not applicable\\
CFG scale
& $3.0$
& Not applicable\\
Initialization / input
& Standard Gaussian noise
& Observed clean motion\\
Motion length
& Reference length, clipped to $[2,300]$ frames at 30 FPS
& Input motion length\\
Response limit / stopping
& Not applicable
& 128 new tokens; stop at the configured EOS token\\
Text sampling
& Not applicable
& Disabled; temperature $0$\\
Prompt
& ``Generate a complete motion from the following description:'' followed by the caption
& ``Describe this motion:'' with the motion input\\
Random seed
& $42+i$, with $i$ the original protocol index
& Deterministic decoding\\
Weights
& EMA
& EMA\\
Evaluation passes
& One in the inspected evaluation records
& One in the inspected evaluation records\\
Output handling
& Denormalize HY201 and convert to the evaluator representation
& Decode up to \texttt{<|im\_end|>}; strip surrounding whitespace\\
\bottomrule
\end{tabularx}
\end{table}

\subsection{Inference Efficiency}
Table~\ref{tab:supp_efficiency} reports single-GPU measurements for both
model sizes with EMA weights and BF16 autocast. T2M uses 50 Euler steps,
CFG 3.0, and separate conditional/unconditional passes. M2T uses greedy
decoding with EOS stopping and a limit of 128 new tokens; actual output
lengths are reported because they affect decoding latency.
The fixed inputs are the first eight captions in the T2M evaluation protocol
and the first eight eligible M2T motions in sorted ID order, cropped to
150 frames. Batch-one measurements use the first input in each set.
Timing includes context encoding and generation, with CUDA synchronization
around each call. Model loading, dataset I/O, motion conversion, smoothing,
and rendering are excluded. These repeated-input measurements characterize
inference cost rather than average latency over the full test set.

\begin{table}[t]
\centering\small
\caption{\textbf{Inference efficiency.} Each input has 150 motion
frames (5 seconds at 30 FPS). Latency is the mean per batch over five
measured calls after two warmup calls. Memory is peak PyTorch allocated
memory, including model weights.}
\label{tab:supp_efficiency}
\begin{tabularx}{\linewidth}{@{}llrrrrX@{}}
\toprule
Model & Task & Batch & Latency (s) & Clips/s & Memory (GiB) & Tokens/clip\\
\midrule
0.6B & T2M & 1 & 4.988 & 0.20 & 1.99 & ---\\
0.6B & T2M & 8 & 6.462 & 1.24 & 2.07 & ---\\
0.6B & M2T & 1 & 0.319 & 3.13 & 2.00 & 10.0\\
0.6B & M2T & 8 & 0.898 & 8.91 & 2.16 & 12.4\\
1.7B & T2M & 1 & 5.072 & 0.20 & 5.91 & ---\\
1.7B & T2M & 8 & 6.547 & 1.22 & 6.00 & ---\\
1.7B & M2T & 1 & 0.321 & 3.11 & 5.92 & 10.0\\
1.7B & M2T & 8 & 0.985 & 8.12 & 6.10 & 14.6\\
\bottomrule
\end{tabularx}
\end{table}

\section{Metrics and Evaluation Protocols}
\label{app:eval}
\subsection{Evaluation Protocols}
\paragraph{Text-to-motion.}
We evaluate on 4,042 selected-caption HumanML3D test cases using the
MotionStreamer MS-272 evaluator~\citep{motionstreamer}. Generated motions
are converted to the evaluator representation, and text--motion retrieval
uses candidate batches of 32.

\paragraph{Motion-to-text.}
We follow the TM2T protocol~\citep{tm2t}, comprising 4,400 cases:
3,969 full motions and 431 temporal subclips. Each case has three reference
captions. Linguistic metrics use the raw reference captions; semantic
retrieval uses the HumanML3D matching network with candidate batches of 32.

\paragraph{Editing and temporal completion.}
Editing follows the MotionLab protocol~\citep{motionlab} on MotionFix,
with the unchanged source included as a no-edit baseline. Temporal completion
covers prediction, backcasting, in-betweening, and keyframe infilling,
corresponding to the observation patterns in Table~\ref{tab:supp_layouts}.
Comparisons use matched sample IDs, observations, target lengths, and text
conditions. Motions are converted to HumanML3D-263 where required by the
evaluator.

\subsection{Evaluation Metrics}
Table~\ref{tab:supp_metrics} summarizes the metrics and their preferred
directions. Embedding-based scores depend on the task-specific evaluator
and should not be compared across different embedding spaces.
For temporal completion, MPJPE and [P]-MPJPE distinguish motion reconstruction
from preservation of observed frames; [P] does not denote Procrustes alignment.

\begin{table}[t]
\centering\small
\renewcommand{\arraystretch}{1.1}
\caption{\textbf{Meaning of the evaluation metrics.} $\uparrow$ and $\downarrow$ indicate
preferred directions; $\rightarrow$ indicates closeness to the real-data value.}
\label{tab:supp_metrics}
\begin{tabularx}{\linewidth}{@{}lX@{}}\toprule
Metric & Definition and interpretation\\\midrule
R@$k$ $\uparrow$ & Fraction of queries whose paired target ranks among the top $k$ candidates in the task-specific embedding space. Ranks are computed within the candidate set of the corresponding protocol.\\
FID $\downarrow$ & Fr\'echet distance between Gaussian fits to generated and real motion embeddings; measures distributional agreement.\\
MMDist / MM-D $\downarrow$ & Mean paired text--motion embedding distance for T2M/M2T; the editing table uses motion--motion distance. This is not maximum mean discrepancy.\\
M2M-Dist $\downarrow$ & Mean distance between generated and target motion embeddings for editing.\\
AvgR $\downarrow$ & Mean rank of the paired target in editing retrieval; depends on candidate-set size.\\
Diversity $\rightarrow$ & Mean embedding distance between sampled motion pairs; compare with the real-data reference.\\
BLEU-1 / BLEU-4 $\uparrow$ & Reference-based n-gram precision with brevity penalty.\\
ROUGE-L $\uparrow$ & Reference-based longest-common-subsequence overlap.\\
CIDEr $\uparrow$ & Consensus-weighted n-gram agreement with reference captions.\\
BERT $\uparrow$ & Contextual-token semantic agreement, reported as rescaled BERTScore F1.\\
MPJPE $\downarrow$ & Mean Euclidean distance between predicted and reference joint positions over the evaluated frames.\\
{[P]}-MPJPE $\downarrow$ & Joint-position error on observed/preserved frames;
here [P] means preservation, not Procrustes alignment.\\
\bottomrule\end{tabularx}
\end{table}

\section{More Results}
\label{app:qualitative}
\subsection{Text-to-Motion Generation}
\begin{figure}[t]
\centering
\includegraphics[width=\linewidth]{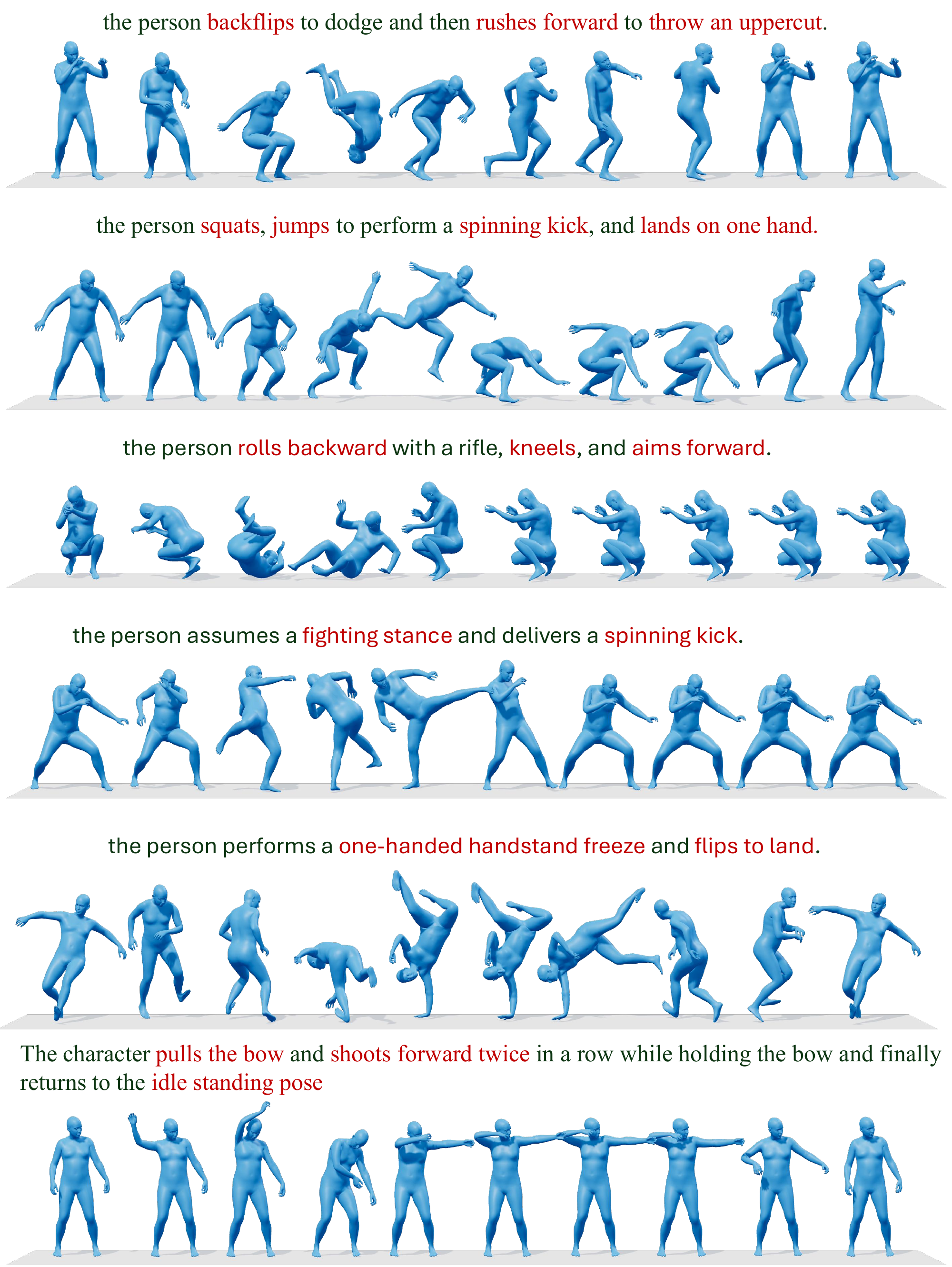}
\caption{\textbf{Additional text-to-motion results.} Motion sequences are shown with
their conditioning descriptions to illustrate prompt adherence.}
\label{fig:supp_t2m}
\end{figure}

Figure~\ref{fig:supp_t2m} presents additional text-to-motion results for the
shown prompts, complementing the qualitative comparisons in the main text.

\subsection{Motion-to-Text Generation}
\begin{figure}[t]
\centering
\includegraphics[width=\linewidth]{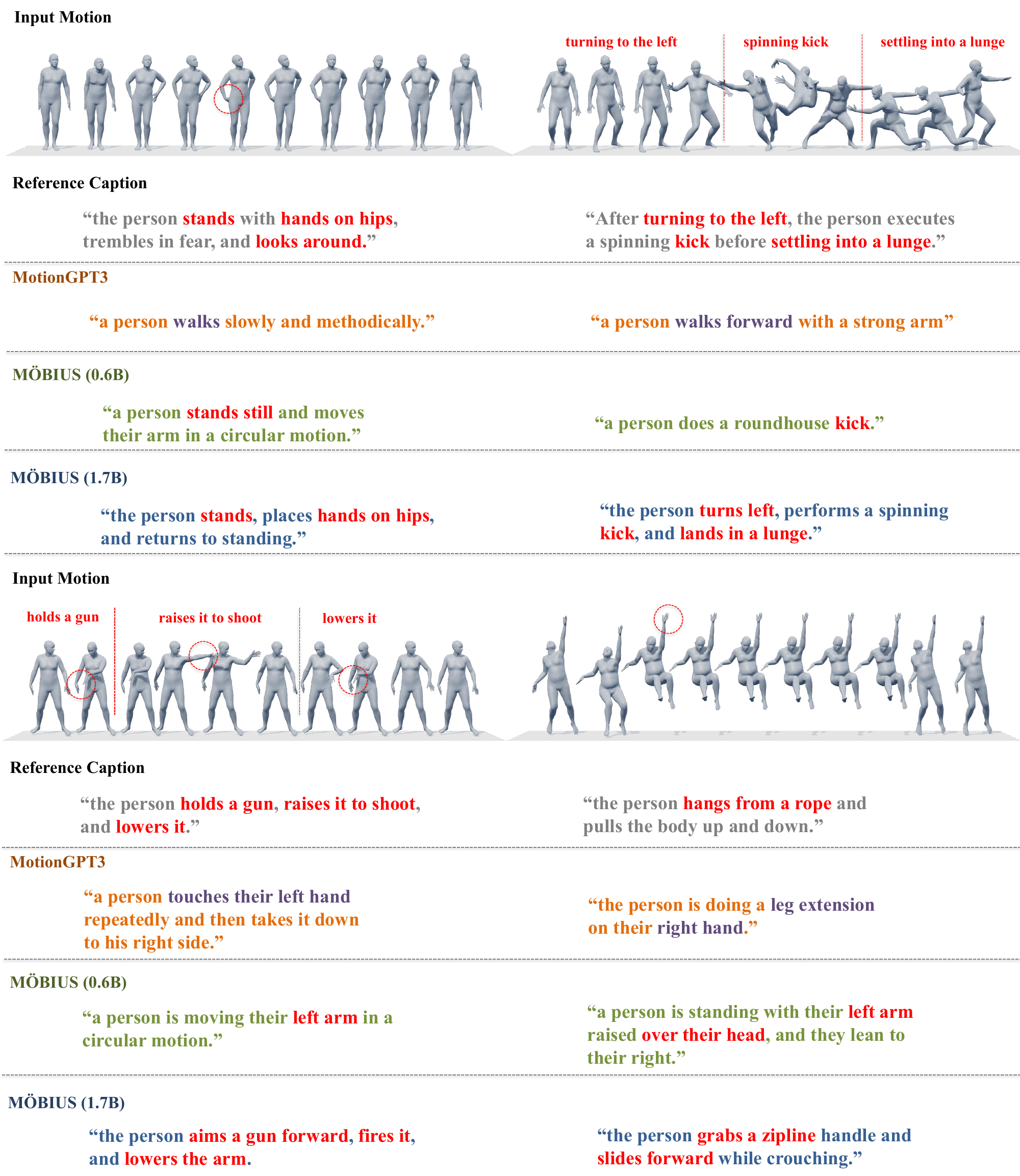}
\caption{\textbf{Additional motion-to-text examples.} The motion sequences and
corresponding captions illustrate motion understanding at the example level.}
\label{fig:supp_m2t}
\end{figure}

Figure~\ref{fig:supp_m2t} presents additional motion-captioning examples,
complementing the quantitative evaluation in the main text.

\subsection{Motion Knowledge Question Answering}
\label{app:capabilities}
Table~\ref{tab:supp_motionqa} presents text-only probes of motion knowledge and
motion-language instruction generation. Base and ours receive the same English
questions without any motion input. These examples probe motion-related knowledge and instruction generation without motion input, complementing the motion-conditioned captioning evaluation.
\begin{table}[t]
\centering\small
\caption{\textbf{Qualitative comparison on motion-knowledge QA and motion-language
instruction generation.} All questions are text-only and require no motion input.}
\label{tab:supp_motionqa}
\renewcommand{\arraystretch}{1.15}
\begin{tabularx}{\linewidth}{@{}lXX@{}}
\toprule
& Base & Ours\\
\midrule

\multicolumn{3}{@{}p{0.97\linewidth}@{}}{
\textbf{Q1.} How does the body's height change when stepping up versus stepping down?
}\\[3pt]
Answer &
When stepping up, the body moves upward as the person lifts the leg and
pushes onto the higher surface. When stepping down, the body moves downward
as the person lowers the foot to the lower surface and transfers their
weight onto it. &
The body's center of mass rises when stepping up and lowers when stepping
down, following the change in support height. \\[6pt]

\midrule

\multicolumn{3}{@{}p{0.97\linewidth}@{}}{
\textbf{Q2.} What happens to the knee and hip joints when a person squats down?
}\\[3pt]
Answer &
As a person squats down, both the knees and hips bend to lower the body.
The knees move forward while the hips move backward, allowing the person
to descend while maintaining balance. &
Both the knees and hips flex as the body lowers, with increasing joint
flexion toward the bottom of the squat. \\[6pt]

\midrule

\multicolumn{3}{@{}p{0.97\linewidth}@{}}{
\textbf{Q3.} Why is it harder to maintain balance when hopping on one foot than when jumping on both feet?
}\\[3pt]
Answer &
Hopping on one foot is more difficult because the body has only one foot
supporting its weight. This provides less stability and requires more
coordination to keep the body upright, whereas using both feet gives a
more stable base of support. &
One-foot hopping has a smaller support area, requiring tighter control of
the center of mass to remain balanced over the supporting foot. \\[6pt]

\midrule

\multicolumn{3}{@{}p{0.97\linewidth}@{}}{
\textbf{Q4.} Generate three random text instructions that could be used for text-to-motion generation.
}\\[3pt]
Answer &
1. Walk forward naturally for a few steps, as if approaching something
in front of you. \newline
2. Try a simple jumping movement where you bend your legs, push yourself
off the ground, and then land again. \newline
3. Slowly turn your body toward the left while moving your arms naturally
with the motion. &
1. A person walks forward three steps and turns to the left. \newline
2. A person crouches down, touches the ground, and stands back up. \newline
3. A person jumps forward and lands with both knees bent. \\[6pt]

\midrule

\multicolumn{3}{@{}p{0.97\linewidth}@{}}{
\textbf{Q5.} What kind of movement is Tai Chi, and how is it typically performed?
}\\[3pt]
Answer &
Tai Chi is a traditional Chinese martial art known for its slow, gentle,
and controlled movements. It is typically performed as a sequence of
flowing postures, with an emphasis on relaxation, balance, breathing,
and coordination. Practitioners move continuously from one posture to
another in a calm and deliberate manner. &
Tai Chi uses slow, continuous movements with smooth weight shifts, gradual
torso rotations, and coordinated curved motions of the arms. \\

\bottomrule
\end{tabularx}
\end{table}

\subsection{Temporal Completion}
Table~\ref{tab:structured-generation} reports results under four observation
patterns. The 0.6B model achieves the lowest MPJPE in all four tasks, while
the 1.7B model achieves the lowest FID throughout. For example, in prediction,
our MPJPE reaches 7.177 compared with the best baseline's 8.781; in keyframe
infilling, our best FID is 0.286 compared with 0.358.
Observed-frame preservation remains task-dependent: HY-Motion has the lowest
[P]-MPJPE for prediction and Kimodo for in-betweening, whereas our models
lead for backcasting and keyframe infilling. These results show complementary
strengths in reconstruction and distributional quality, rather than a uniform
advantage on every metric. The two model sizes also use different training
data, so their comparison does not isolate model scale.

\paragraph{Effect of multi-task training on T2M and M2T.}
Table~\ref{tab:supp_multitask_transfer} compares the main-table 0.6B model
with the multi-task variant after 20k continuation steps. Adding conditional
generation tasks yields a modest T2M trade-off: R@1 decreases by 1.53
percentage points, FID increases from 6.1291 to 6.6746, and MMDist increases
from 13.9973 to 14.0824. Nevertheless, the multi-task model still outperforms
all other HumanML3D-trained methods in Table~\ref{tab:t2m} on R@1--3, FID,
and MMDist, including generation-specialized models. Its M2T scores remain
identical to the main-table results, consistent with keeping the understanding
parameters frozen during the additional generation training.

\paragraph{Why report two training variants?}
The main-table comparison focuses on the bidirectional T2M/M2T setting shared
with unified motion--language baselines, reducing the influence of additional
task-specific supervision on that comparison. The multi-task evaluation asks
a complementary question: can the same architecture accommodate additional
observation patterns and conditional generation objectives while retaining
its core capabilities? The temporal-completion results, together with the
strong T2M scores and unchanged M2T performance, support this broader task
coverage without introducing task-specific generation modules. We therefore
report the variants separately to make the trade-off between task breadth
and T2M specialization explicit. This demonstrates extensibility across
trained tasks, rather than establishing zero-shot generalization to unseen
tasks or distributions.

\begin{table}[t]
\centering\small
\caption{\textbf{T2M performance of the main-table and multi-task 0.6B variants.}
The first row reproduces Table~\ref{tab:t2m}; the second uses the multi-task
20k EMA checkpoint. FID is reported in the evaluator's raw embedding space,
as in the main table. M2T scores are unchanged.}
\label{tab:supp_multitask_transfer}
\begin{tabularx}{\linewidth}{@{}l*{5}{>{\centering\arraybackslash}X}@{}}
\toprule
Variant & R@1$\uparrow$ & R@2$\uparrow$ & R@3$\uparrow$ & FID$\downarrow$ & MMDist$\downarrow$\\
\midrule
Main-table model & 0.8209 & 0.9397 & 0.9670 & 6.1291 & 13.9973\\
Multi-task model & 0.8056 & 0.9221 & 0.9583 & 6.6746 & 14.0824\\
\bottomrule
\end{tabularx}
\end{table}

\begin{table}[t]
\caption{Motion temporal completion on HumanML3D.
[P]-MPJPE measures the preservation error on conditioned frames.
Lower MPJPE, [P]-MPJPE, FID, and MM-D are better, while higher
R-Precision is better.}
\label{tab:structured-generation}
\centering
\scriptsize
\setlength{\tabcolsep}{3.1pt}
\begin{tabularx}{\linewidth}{@{}l*{7}{>{\centering\arraybackslash}X}@{}}
\toprule
Method
& MPJPE$\downarrow$
& [P]-MPJPE$\downarrow$
& FID$\downarrow$
& R\@1$\uparrow$
& R\@2$\uparrow$
& R\@3$\uparrow$
& MM-D$\downarrow$ \\
\midrule

GT
& 0.000 & 0.000 & 0.000 & 0.448 & 0.646 & 0.767 & 3.466 \\
\midrule

\multicolumn{8}{c}{\emph{Motion Prediction}} \\
\midrule
CondMDI~\citep{condmdi}
& 11.439 & 5.882 & 3.213 & 0.383 & 0.565 & 0.667 & 3.927 \\
Kimodo~\citep{kimodo}
& 8.781 & 1.368 & 1.480 & 0.337 & 0.503 & 0.616 & 4.532 \\
HY-Motion (1B)~\citep{hymotion}
& 9.372 & \textbf{0.856} & 3.185 & 0.401 & 0.592 & 0.701 & 4.148 \\
MotionLab~\citep{motionlab}
& 9.829 & 4.086 & 1.108 & 0.436 & 0.626 & 0.733 & 3.479 \\
Ours (0.6B)
& \textbf{7.177} & \underline{1.232} & \underline{0.543}
& \underline{0.493} & \textbf{0.696} & \textbf{0.806} & \textbf{3.099} \\
Ours (1.7B)
& \underline{7.754} & 1.320 & \textbf{0.435}
& \textbf{0.498} & \underline{0.695} & \underline{0.802} & \underline{3.118} \\

\midrule
\multicolumn{8}{c}{\emph{Motion Backcasting}} \\
\midrule
CondMDI~\citep{condmdi}
& 11.704 & 8.049 & 3.895 & 0.361 & 0.553 & 0.656 & 4.054 \\
Kimodo~\citep{kimodo}
& 8.794 & \underline{2.408} & 0.969 & 0.345 & 0.532 & 0.649 & 4.301 \\
HY-Motion (1B)~\citep{hymotion}
& 9.118 & 2.857 & 1.344 & 0.426 & 0.617 & 0.726 & 3.848 \\
MotionLab~\citep{motionlab}
& 9.923 & 5.438 & 2.292 & 0.474 & 0.682 & 0.785 & \textbf{3.123} \\
Ours (0.6B)
& \textbf{6.812} & \textbf{2.291} & \underline{0.519}
& \textbf{0.483} & \underline{0.685} & \textbf{0.801} & 3.135 \\
Ours (1.7B)
& \underline{7.419} & 2.459 & \textbf{0.416}
& \underline{0.482} & \textbf{0.689} & \underline{0.799} & \underline{3.125} \\

\midrule
\multicolumn{8}{c}{\emph{Motion In-Betweening}} \\
\midrule
CondMDI~\citep{condmdi}
& 11.299 & 6.869 & 2.878 & 0.387 & 0.568 & 0.676 & 3.914 \\
Kimodo~\citep{kimodo}
& 7.604 & \textbf{1.436} & 1.251 & 0.353 & 0.532 & 0.655 & 4.347 \\
HY-Motion (1B)~\citep{hymotion}
& 9.457 & 1.986 & 6.183 & 0.358 & 0.541 & 0.651 & 4.679 \\
MotionLab~\citep{motionlab}
& 8.984 & 4.422 & 1.213 & 0.434 & 0.631 & 0.740 & 3.474 \\
Ours (0.6B)
& \textbf{6.354} & \underline{1.655} & \underline{0.465}
& \underline{0.488} & \underline{0.684} & \textbf{0.803} & \textbf{3.131} \\
Ours (1.7B)
& \underline{6.724} & 1.666 & \textbf{0.371}
& \textbf{0.489} & \textbf{0.685} & \underline{0.802} & \underline{3.145} \\

\midrule
\multicolumn{8}{c}{\emph{Keyframe Infilling}} \\
\midrule
CondMDI~\citep{condmdi}
& 7.324 & 6.711 & 2.881
& 0.442 & 0.545 & 0.742 & 3.694 \\
Kimodo~\citep{kimodo}
& 5.015 & 2.485 & \underline{0.358}
& 0.397 & 0.606 & 0.727 & 3.848 \\
HY-Motion (1B)~\citep{hymotion}
& 11.984 & 7.314 & 18.878
& 0.275 & 0.431 & 0.534 & 6.016 \\
MotionLab~\citep{motionlab}
& 5.954 & 4.262 & 1.225
& 0.478 & 0.682 & 0.792 & \textbf{3.136} \\
Ours (0.6B)
& \textbf{4.501} & \underline{2.394} & 0.398
& \textbf{0.492} & \textbf{0.699} & \textbf{0.811} & \underline{3.208} \\
Ours (1.7B)
& \underline{4.541} & \textbf{2.333} & \textbf{0.286}
& \underline{0.489} & \underline{0.691} & \underline{0.803} & 3.278 \\
\bottomrule
\end{tabularx}
\end{table}

\end{document}